\documentclass[letterpaper, 10 pt, conference]{ieeeconf}  % Comment this line out if you need a4paper
\IEEEoverridecommandlockouts                              % This command is only needed if
\usepackage{epsfig} % for postscript graphics files
\usepackage{times} % assumes new font selection scheme installed
\usepackage{hyperref}
\usepackage{mathtools}
\usepackage{xcolor}

\usepackage{amsmath} % assumes amsmath package installed
\usepackage{amssymb}  % assumes amsmath package installed
\usepackage{mathrsfs}
\usepackage{comment} % assumes amsmath package installed
\usepackage{algorithm,algorithmic}

\title{\Large \bf
Global Position Control on Underactuated Bipedal Robots:\\
Step-to-step Dynamics Approximation for Step Planning%\author{Xiaobin Xiong$^{1}$, Aaron D. Ames$^{2}$% %%%%  non-periodic walking on underactuated bipedal robots. <-this % stops a space
   \author{Xiaobin Xiong, Jenna Reher and Aaron D. Ames% <-this % stops a space
\thanks{This work was supported by NSF 1924526 and 1923239.}
\thanks{The authors are with the Department of Mechanical and Civil Engineering, California Institute of Technology, Pasadena, CA 91125
       { Contact: \tt\small xxiong@caltech.edu}}%
 }}
\begin{document}
\maketitle
\thispagestyle{empty}
\pagestyle{empty}

\begin{abstract}
Global position control for underactuated bipedal walking is a challenging problem due to the lack of actuation on the feet of the robots. In this paper, we apply the Hybrid-Linear Inverted Pendulum (H-LIP) based stepping on 3D underactuated bipedal robots for global position control. The step-to-step (S2S) dynamics of the H-LIP walking approximates the actual S2S dynamics of the walking of the robot, where the step size is considered as the input. Thus the feedback controller based on the H-LIP approximately controls the robot to behave like the H-LIP, the differences between which stay in an error invariant set. Model Predictive Control (MPC) is applied to the H-LIP for global position control in 3D. The H-LIP stepping then generates desired step sizes for the robot to track. Moreover, turning behavior is integrated with the step planning. The proposed framework is verified on the 3D underactuated bipedal robot Cassie in simulation together with a proof-of-concept experiment.
\end{abstract}

\IEEEpeerreviewmaketitle

%\citet{Supplementary}
\section{INTRODUCTION}

%%%% path tracking is an important functionality of bipedal robots; typically solution zMP, MPC ....
Global position control (e.g, following a path) is an important behavior for enabling bipedal robots and humanoids to locomote through cluttered environments \cite{clary2018monte, agrawal2017discrete}. It has been realized on fully actuated humanoid robots, typically through a model predictive control approach \cite{scianca2019mpc,apgar2018fast,4115592,8794117, naveau2016reactive, fallon2015architecture} for planning the footsteps and center of mass (COM) trajectories \cite{tanguy2019closed} with the constraint that the Zero Moment Point (ZMP) always remains inside the support polygon \cite{vukobratovic2004zero}. The desired COM dynamics can be embedded on full-dimensional robots \cite{kajita2003biped, 8815144, 8461140}. However, fully actuated robots using ZMP controllers oftentimes lead to slow and conservative walking motions due to the relatively large inertias of the legs and limited actuation on the ankles.

%%%% the foot actuation doesnot result in dynamic walking, however path tracking is difficult on underactuated robots
Underactuated bipedal robots, where feet are not actuated and thus light, typically can produce more dynamic walking behaviors \cite{rezazadeh2015spring, raibert1986legged, SreenathPPG11}. The downside of the underactuation lies in the difficulty of balancing. It requires appropriate stepping to avoid falling \cite{da20162d,xin2019online, hodgins1991adjusting,kim2020dynamic}. The notion of stability is not well-understood in contrast to foot stability with ZMP for fully actuated robots. Because of this, the literature has been largely focused on generating and stabilizing local periodic behaviors of 3D underactuated bipedal walking robots \cite{westervelt2003hybrid, garcia1998simplest, powell2015model, chen2020optimal} or utilizing existing stable walking for certain tasks \cite{nguyen2017dynamic}. Global position control on underactuated walking has been much less explored.

 \begin{figure}[t]
      \centering
      \includegraphics[width = 0.8\columnwidth]{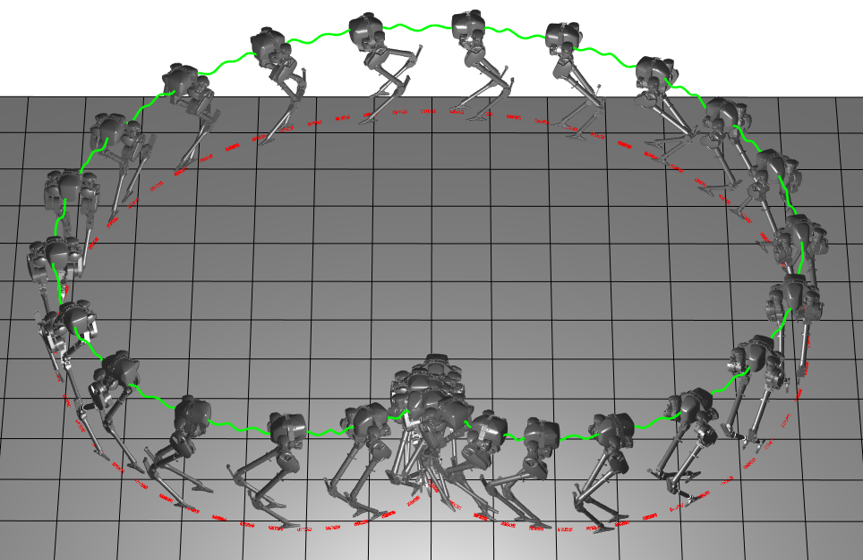}%  inAir.jpg}
      \caption{An example of controlling the global position of the robot to follow a cardioid path.}
      \label{torsoHLIP}
\end{figure}
%%%% contributtions methods

To address the global position control problem on underactuated bipeds, we apply the step planning based on the Hybrid Linear Inverted Pendulum (H-LIP) model \cite{xiong20213d, xiong2020ral}. The H-LIP approximates the underactuated translational dynamics of the walking, and the step-to-step (S2S) dynamics of the H-LIP can also approximate the S2S dynamics of the underactuated robotic walking. The Model Predictive Control (MPC) is formulated on the H-LIP for realizing a global walking behavior; the H-LIP based stepping is then applied on the robot to keep the error between the horizontal states of the robot and the states of the H-LIP in an error (disturbance) invariant set. Thus the global walking behavior is approximately realized on the robot.

 %\textbf{Minimum computation. computation friendly.  No cop manipulation}

%%% on how to generate periodic walking in the first place
 %Walking on desired step locations from planning can be realized via perturbing an existing periodic stepping-in-place motion via changing the desired step sizes. The generation of the stepping-in-place motion is similar to \cite{xiong20213d}. The desired output trajectories are tracked by control Lyapunov function based Quadratic programs (CLF-QPs) \cite{ames2014rapidly, XiongSLIP}. Additionally, by assuming the transversal motion and the translational motion are weakly coupled, turning is individually planned and then integrated with the footstep planning.

%Similar to previous work \cite{xiong20213d} on generating periodic walking, we use an actuated spring-loaded inverted pendulum (aSLIP) model to optimize a periodic leg length trajectories for the 3D underactuated robot, which are used as desired trajectories to track. The desired swing positions are constructed smoothly based on the desired step locations. The desired trajectories are tracked by control Lyapunov function based Quadratic programs (CLF-QPs) \cite{ames2014rapidly, XiongSLIP}.

The use of the H-LIP is different when compared to the application of the canonical LIP on \textit{fully actuated humanoids} \cite{scianca2019mpc, naveau2016reactive,tanguy2019closed, krause2012stabilization}. The foot underactuation prevents directly controlling the robot's COM dynamics to follow pendulum models exactly, since there is no ankle torque and the natural dynamics of the two are different. The perspective of the approach in this paper is to view the model difference between the underactuated robot and the H-LIP as disturbances to the H-LIP, and the feedback controller is applied to bound the error between two systems.

The proposed approach of the step planning via H-LIP for global position control is implemented on the underactuated 3D biped Cassie in simulation for tracking various paths with different shapes, including: a circle, a sinusoid, a square, and a cardioid. Moreover, the method can also avoid obstacles and reject disturbances during walking. Additionally, as a proof-of-concept, we experimentally verify the approach by tracking one simulated walking path showing dynamic consistency, i.e., that the path produced by the robot in simulation can be tracked on the robot hardware. 
The goal is to enable underactuated bipedal robots to walk under global position control in cluttered and dynamic environment autonomously.
%e, it is the first holistic approach to global position control on 3D underactuated bipedal robots.

%An idea on how to discuss the experimental results: 
%Additionally, as a proof-of-concept, we experimentally verify the approach by tracking the simulated walking paths showing dynamic consistency, i.e., that the paths produced by the robot in simulation via the H-LIP can be tracked on the full-order hardware. Note: will need to work on how to frame the experimental results.  The above is one idea, which I expounded upon in my email. 
\begin{comment}
\cite{bhounsule2014low} discrete LQR controls on heuristically selected quantities to stabilize designed trajectories, locally linearized return map.
\cite{martin2017experimental} experimentally validated a deadbeat controller on SLIP running on the ATRIAS robot in planar setting.
\end{comment}

% This work

\section{Global Position Control via Stepping}
\label{overview}
In this section, we describe the problem which we are interested to solve and explain the approach of the stepping based on the Hybrid-Linear Inverted Pendulum (H-LIP). 
\subsection{Problem Definition}
We consider the bipedal robot walks on flat ground in a 3D environment. The robot is given a walking path with a terminal location that it should reach. We assume the path is planned via a high-level planner from all the sensors on the robot. The path is supposed to be relatively-smooth and obstacle-free. Additionally, the path is also generated with a speed profile, which is assumed to be feasible for the robot to realize. We parameterize the desired path by $\mathbf{r}(t)$, which is:
\begin{equation}
\mathbf{r}(t) = [ r_x^d(t), r_y^d(t), r_\theta^d(t) ]^T,
\end{equation} 
where $r_x^d(t), r_y^d(t)$ are the positions in the global frame and $r_\theta^d(t)$ be the angle of the tangent line to the path. The task for the walking is to drive the robot (depicted by its COM) to follow the path with a given time. 

Note that, the provided path is not necessary for global position control. The approach in the paper can also handle the cases when the robot has to avoid obstacles by itself via the global position control. This becomes useful if the robot moves in a dynamic environment where an obstacle-free path may be difficult to plan in real-time. Additionally, the approach can handle unknown external disturbances. These applications will be presented in the results.

\subsection{H-LIP Based Stepping via S2S Approximation}
%%%%%%%%%%%%%%%%%%%%%%%%%%%%%%%%%%%%%%%%%%%%%%%%%%%%%%%%%%%%%%%%%%%%%%%%%%%%%%%%%%%%%%%%%%%%%%%%%%%%%%%%%%%%%%%%
Direct control of the COM of the robot to follow a trajectory is not possible on \textit{underactuated} bipedal walking. Instead, controlling the location of the foot (foot-placement) to indirectly manipulate its COM behavior is possible. In the literature, foot-placement has been mostly studied for stabilizing (local) walking behaviors, e.g., controlling the desired velocity on Raibert hoppers \cite{raibert1986legged}. 

\begin{figure}[t]
      \centering
      \includegraphics[width = 1\columnwidth]{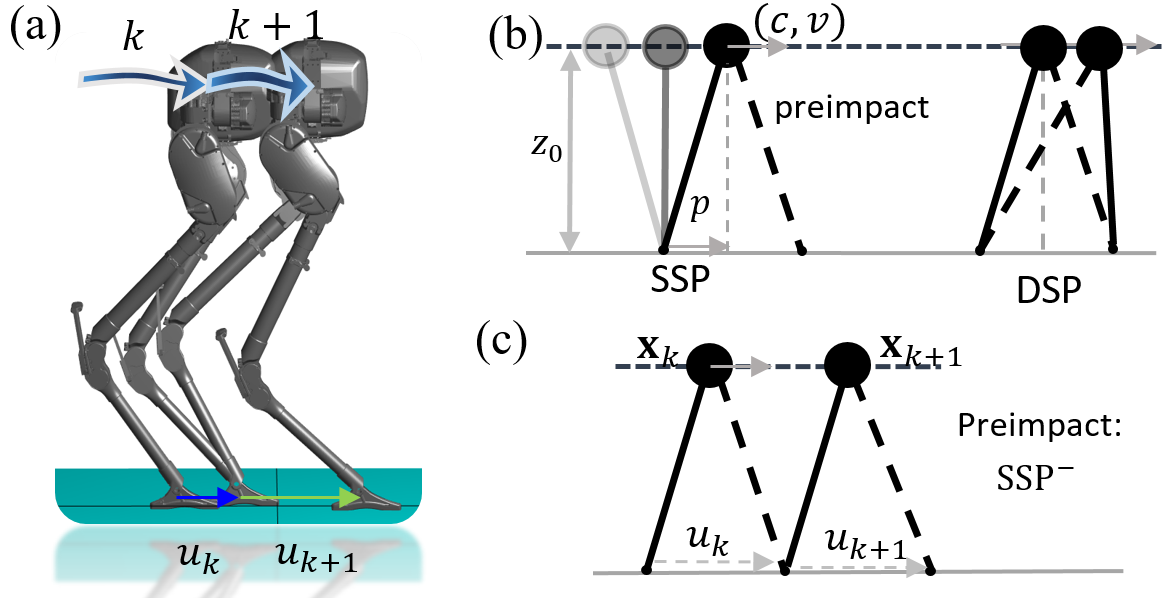}
      \caption{(a) The step-to-step (S2S) dynamics illustrated on the robot Cassie. (b) The Hybrid-Linear Inverted Inverted Pendulum (H-LIP) model. (c) The S2S dynamics of the H-LIP. }
      \label{fig:HLIP}
\end{figure}

The goal is to control the \textit{global} position of the COM while stabilizing the walking to prevent falling. We apply the H-LIP based stepping \cite{xiong20213d, xiong2020ral} to solve this problem.
It treats the step size as the input to the discrete step-to-step (S2S) dynamics of the horizontal state at the impact event (Fig. \ref{fig:HLIP} (a)). To illustrate this, we assume an existing periodic walking, e.g., a stepping-in-place. The pre-impact state undergoes a discrete S2S dynamics at the step level: 
\begin{equation}
\label{eq:robotS2S}
    \{q, \dot{q}\}^{-}_{k+1} = \mathcal{P}( \{q, \dot{q}\}^{-}_k, \tau(t)),
\end{equation}
where $q$ is the configuration of the robot, $\{q, \dot{q}\}^{-}$ indicates the pre-impact state at each step, and $\tau$ is the torque applied during this step. During walking, we care about the \textit{horizontal pre-impact state} of the COM of the robot, which is denoted by $\mathbf{x}^h = [c^{x},p^{x}, v^{x}]^T$ in the $x-z$ plane and $\mathbf{y}^h = [c^{y},p^{y}, v^{y}]^T$ in the $y-z$ plane. ${c}^{x, y}$ are the horizontal position of the COM in the inertial frame, ${p}^{x, y}$ are its horizontal position relative to its stance foot, and ${v}^{x, y}$ are the horizontal velocity of the COM. Thus the S2S dynamics of the horizontal state can be presented by:
\begin{align}
\label{eq:s2s_xz}
    \mathbf{x}^h_{k+1} = & \mathcal{P}^h_\mathbf{x}( \{q, \dot{q}\}^{-}_k, \tau(t)), \\
 \label{eq:s2s_yz}
   \mathbf{y}^h_{k+1} = & \mathcal{P}^h_\mathbf{y}( \{q, \dot{q}\}^{-}_k, \tau(t)).
\end{align}

The actual S2S is very complex and cannot be calculated in closed form. We use the S2S dynamics of the H-LIP (Fig. \ref{fig:HLIP} (c)) to approximate the horizontal S2S of the robot in each plane. For instance, Eq. \eqref{eq:s2s_xz} can be rewritten as: 
\begin{align}
    \mathbf{x}^h_{k+1} &= A \mathbf{x}^h_k + B u^x_k + w \\
    \label{eq:w}
    w:&= P^h_\mathbf{x} - A \mathbf{x}^h_k - B  u^x_k
\end{align}
where $w \in W$ is treated as the disturbance to the linear dynamics, and $u^x$ is the step length on the robot in the $x-z$ plane. $A$ and $B$ are constant matrices which come from the S2S of the H-LIP \cite{xiong2020ral, xiong2021slip}:
\begin{equation}
\label{eq:HLIP_S2S}
    \mathbf{x}^{\text{H-LIP}}_{k+1} = A \mathbf{x}^{\text{H-LIP}}_{k} + B u_k^{x^\text{H-LIP}} 
\end{equation}
where $u^{x^\text{H-LIP}}$ is the step size on the H-LIP. Exact expressions of the S2S can be found in \cite{xiong2020ral}. $\mathbf{x}^{\text{H-LIP}} = [c^\text{H-LIP}, p^\text{H-LIP}, v^\text{H-LIP}]^T$ is the pre-impact state of the H-LIP. 

Based on the approximation, we apply the \textit{H-LIP based stepping} to generate desired step length for the robot: 
\begin{equation}
\label{eq:Hlip-stepping}
    u^x_k = u^{x^\text{H-LIP}}_k + K(\mathbf{x}^h_k - \mathbf{x}^{\text{H-LIP}}_{k}),
\end{equation}
then the error between the two models $\mathbf{e}^x_k = \mathbf{x}^h_k - \mathbf{x}^{\text{H-LIP}}_{k}$ evolves on the closed-loop error dynamics: 
\begin{equation}
    \mathbf{e}^x_{k+1} = (A+BK) \mathbf{e}^x_{k} + w.
\end{equation}
If $A+BK$ is stable, then the error will converge to an error (disturbance) invariant set $E$. If $\mathbf{e}_k \in E$, then $\mathbf{e}_{k+1} \in E$. In other words, one can control the robot by first controlling the global position of the H-LIP via the linear dynamics and then applying Eq. \eqref{eq:Hlip-stepping} to make the robot close to the H-LIP. 

Therefore, we use the S2S approximation in both the $x-z$ plane and the $y-z$ plane of walking to plan desired step sizes for global position control. Note that the S2S approximation is valid when the walking of the robot is close to that of the H-LIP. In the next two sections, we will explain the control on the H-LIP and then the walking construction of the stepping on the 3D robot.

% Executing the step plan is perturbing the existing walking so that the global translational states of the robot behave to achieve certain behaviors, e.g., following a path. 

%We first present technical definitions of the partial return map, i.e. the step-to-step translational dynamics of the walking systems, for which the step sizes are the inputs/actuations. Then we illustrate the application of the control on the global translational states via an approximation to the actual step-to-step dynamics.

\section{3D Motion Generation on H-LIP}
In this section, we briefly present the H-LIP model. Then we describe the Model Predictive Control (MPC) that is applied on the 3D-H-LIP with an eye towards the application on 3D bipedal robots. The superscript $^{\text{H-LIP}}$
will be omitted in the equations in this section. 
\subsection{Dynamics of the H-LIP}
The H-LIP \cite{xiong20213d, xiong2020ral} is a variant of the canonical LIP model \cite{kajita2003biped}. Compared to the canonical LIP, it has no ankle actuation, but has a double support phase (DSP). In the single support phase (SSP), the H-LIP is a passive LIP model. The duration of each domain is assumed to be constant, and the transition between domains is smooth. The dynamics are: 
\begin{equation}
    \text{SSP}: \ddot{p} = \lambda^2 p, \quad \text{DSP}: \ddot{p} = 0,
\end{equation}
where $\lambda = \sqrt{\frac{g}{z_0}}$ and $z_0$ is the nominal height of the mass. Since the dynamics in both domains are linear, the step-to-step (S2S) dynamics is also linear, as shown in Eq. \eqref{eq:HLIP_S2S}. The exact expressions of the S2S can be found in \cite{xiong2020ral}.

Note that here the H-LIP is used as an approximation to the walking dynamics of underactuated robots, instead of a template system for the fully-actuated robot to embed \cite{ faraji2014robust,villa2017model, griffin2018straight, krause2012stabilization}. While the assumptions of the constant COM height and smooth impact of the foot-ground contacts do not exactly match with the walking of the robot, we will show that this approximation can be utilized to realize approximate walking behaviors on the robot.
%%%%%%%%%%%%%%%%%%%%%%%%%%%
%It is not possible to exact track given path for underactuated walking robots. In the MPC formulation, we will encode the tracking task in the cost function. Dynamic constraints, step size
%The key assumption of the rMPC with turning beheavior is that the sagittal motion, lateral motion and the transveral motion of walking can be decoupled. Similar decoupling ideas have been widely applied in the literature \cite{raibert}.

 %The MPC approach for control naturally inspires a formulation of solving path tracking problem. It is assumed that a high-level planner provides a feasible/optimal path for the system to walk through. We can apply the MPC approach to online generate optimal step location for best-tracking the desired path. For underactuated robotic walking, it is not possible to exactly tracking the given path. So the approach is to set the tracking task as the cost in the MPC while having other bounded inputs and linear dynamics as the constraints.

\subsection{Model Predictive Control on 3D-H-LIP}
The H-LIP is a planar model. For 3D bipedal locomotion, we use the 3D version of the H-LIP, which is an orthogonal composition of two planar H-LIP models. Since the dynamics in each plane are decoupled, the S2S dynamics of the 3D-H-LIP has two linear S2S dynamics, as shown in Eq. \eqref{eq:HLIP_S2S}. 

Additionally, it is desirable to define orientation on the 3D-H-LIP with an eye towards enabling turning on the 3D robot. With turning, the robot can `face' towards the direction of walking for observing and planning the path using its sensors if necessary. Moreover, the robot typically has different ranges of motion in its sagittal and lateral planes. With turning, the robot can flexibly align its plane with a larger range of motion to the direction of walking. Therefore, we add a trivial torso with no inertia on the point-mass (Fig. \ref{torsoHLIP} (a)) to define the orientation of the H-LIP. The transversal dynamics is trivial and does not affect the translational dynamics.

We apply model predictive control (MPC) on the 3D-H-LIP to control its global position. The MPC is designed to control the point mass to best track a given trajectory on the ground. It is realized via quadratic programs, each of which is solved at every step to find a sequence of optimal step sizes in the next $N$ steps. The first step size in the sequence is applied. The procedure repeats for each step.
%For the requirement of global position tracking, the states of the H-LIP need to be augmented to include the global position of the mass.

\textbf{Optimization Variables:} At each step, the next $N$ steps are planned. Thus, all the pre-impact states $\mathbf{x}_k$ (in the $x-z$ plane), $\mathbf{y}_k$ (in the $y-z$ plane) and all corresponding inputs $u^x_k, u^y_k$ are selected as the optimization variables, where $k = 1, \dots, N$. The current step is indexed as 1.

\textbf{Cost Function:} The cost function of the MPC includes two parts: one is encoding the tracking performance as the distance between the mass states and the desired path, and the other past is penalizing the step sizes. Let $T_{\sum} =T_{\text{SSP}}+T_{\text{DSP}}$ be the period of the walking. Then the look-ahead time horizon is $NT_{\sum}$. The cost function on the tracking performance is defined as:
\begin{equation}
    J_t =\textstyle \sum_{k= 1}^N |\mathbf{c}_k - \mathbf{r}_{x,y}^d(t_k)|^2 +  |\mathbf{v}_k - \dot{\mathbf{r}}_{x,y}^d(t_k)|^2,
\end{equation}
where $\mathbf{c}_k = [c^x_k, c^y_k]^T, \mathbf{v}_k = [v^x_k, v^y_k]^T,\mathbf{r}_{x,y}^d = [r_x^d, r_y^d]^T, t_k = t_0 + k T_{\sum}$ and $t_0$ is the time when the MPC is solved.
Additionally, we add cost on the input to penalize large step sizes,
\begin{equation}
    J_u =\textstyle \sum_{k= 1}^N |\mathbf{u}^{x,y}_k|^2,
\end{equation}
where $\mathbf{u}^{x,y}_k = [u^x_k, u^y_k]^T$.
The final cost function is a combination of the two:
\begin{equation}
    J_{\text{MPC}} = J_{t} + \alpha J_u,
\end{equation}
which is a quadratic function of all the variables. $\alpha \in \mathbb{R}$ is an coefficient to leverage the tracking and planned step sizes.

\textbf{Constraints:} The dynamics in each step are encoded via linear equality constraints as in Eq. \eqref{eq:HLIP_S2S}. Additional constraints include the initial state constraint, step size limits (input limits). The step size limits come from the physical kinematic feasibilities of the robot, which are different in the sagittal and lateral planes.  %Suppose the translational dynamics is always expressed in the inertial frame for the consistency of global position tracking, the turning motion affects available step size in the $x-z$ plane and $y-z$ plane. 
Assuming the torso is controlled to point to the direction of walking, the sagittal plane is aligned with the tangent line (denoted by $r_\theta^d(t)$). The step length in the sagittal plane $s^l$ and the step width $s^w$ in the lateral plane can be expressed as:
\begin{align}
s^l (u^x, u^y) =  u^x \mathrm{cos}(r_\theta^d) + u^y \mathrm{sin}(r_\theta^d), \\
s^w (u^x, u^y)  =  - u^x \mathrm{sin}(r_\theta^d) + u^y \mathrm{cos}(r_\theta^d).
\end{align}
Then the step size constraints are:
\begin{align}
s^l_{\text{min}}  \leq s^l (u^x, u^y) \leq s^l_{\text{max}}, \\
s^w_{\text{min}} \leq s^w (u^x, u^y) \leq s^w_{\text{max}},
\end{align}
where $s^l_{\text{min/max}}, s^w_{\text{min/max}}$ are the available step sizes in each plane, which are linear functions of the states.
Additionally, the robot should avoid kinematic conflicts for foot stepping. It can be easily specified through enforcing a finite minimum step width in the lateral plane, which can be added into the above constraint by changing $s^w_{\text{min/max}}$. 

%The step size difference is that, and the step size difference constraint, which are constraints considered for the application on the robots. $| u_{k+1} - u_{k} | \leq \Delta u $
% where $\Delta u$ is a constant. This constraint avoids the system to dramatically change step sizes consecutively, which may lead to undesirable behaviors on the robot.

\textbf{MPC Formulation:} We compactly present the MPC formulation for the 3D-H-LIP. At each step, a constrained quadratic program (QP) is formulated and solved. The QP is as follows,
\begin{align}
\label{eq:MPC-LIP}
\{\mathbf{u}^{x,y}, \mathbf{x, y}\}_\mathbb{K} =   &\underset{\{\mathbf{u}^{x,y}, \mathbf{x, y}\}_\mathbb{K} \in \mathbb{R}^{8 \times N}    }
 {\text{argmin}} J_{\text{MPC}} \\
\text{s.t.} \quad & \mathbf{x}_{k+1} = A \mathbf{x}_k + B u^x_k \nonumber \tag{S2S}\\
                 & \mathbf{y}_{k+1} = A \mathbf{y}_k + B u^y_k \nonumber \tag{S2S}\\
                 &  s^{l,w}_{\text{min}} \leq s^{l, w} (u^x_k, u^y_k) \leq  s^{l,w}_{\text{max}} \nonumber \tag{Input}
% \quad &  \left |  u_k \right | < u_{\text{max}} , k = \{1, \dots, N\}  \nonumber \\
% \quad &  | u_{k+1} - u_{k} | \leq \Delta U , k = \{1, \dots,  N-1\}   \nonumber \\
 %  \quad &  \tilde{\mathbf{x}}_1 = \tilde{\mathbf{x}}_{\text{now}} \  \nonumber
\end{align}
where $\forall k \in \mathbb{K}$, and $\mathbb{K} = \{1, \dots, N\}$.
% where $\tilde{\mathbf{x}}_{\text{now}}$ is the final state of current step of the H-LIP. 
The first solution of the step size $\mathbf{u}^{x,y}_{k=1}$ is applied at current optimization.
\begin{figure}[t]
      \centering
      \includegraphics[width = 1\columnwidth]{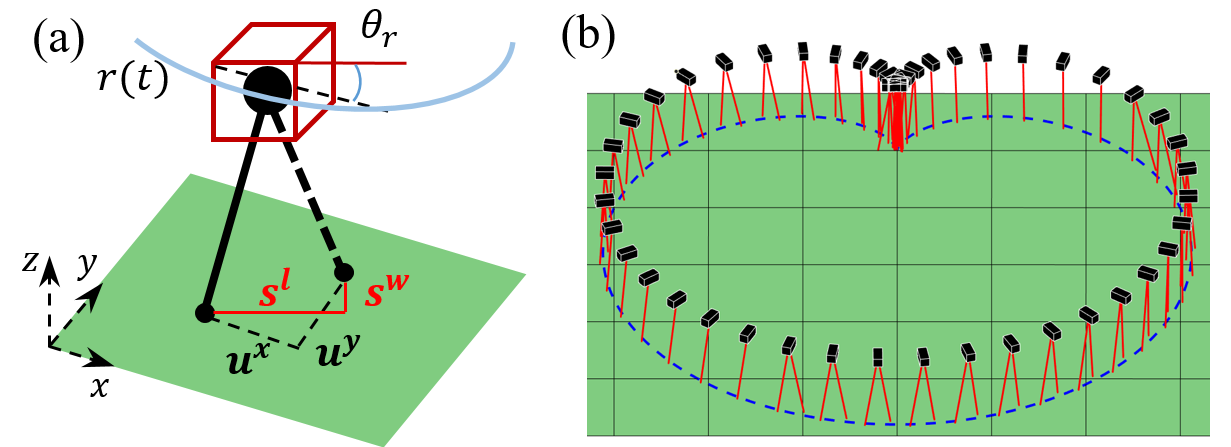}%  inAir.jpg}
      \caption{(a) The 3D-H-LIP with a trivial torso. (b) An example of path tracking of the 3D-H-LIP (blue dashed line is the desired trajectory).}
      \label{torsoHLIP}
\end{figure}
Fig. \ref{torsoHLIP} (b) shows an example of the optimized walking of the 3D-H-LIP in 3D for tracking a cardioid trajectory.
Here we assume that the torso orientation is aligned with the path direction but it can be freely decided. % The control of the orientations of the torso and the feet on the robot are explained in the next section.

\begin{comment}
\textbf{RMPC}
\begin{align}
\label{MPC-LIP}
u_k= & \underset{ \{u_{1, \dots, N}, \tilde{X}_{1, \dots, N} \} \in \mathbb{R}^{N \times 3N}    }
 {\text{argmin}}  J_{\text{MPC}} \\
\text{s.t.} \quad & \tilde{X}_{k+1} = \tilde{A} \tilde{X}_k + \tilde{B} u_k , k = \{1, \dots, N-1\}\nonumber\\
  \quad &  \tilde{X}_1 = \tilde{X}_{\text{now}} \oplus E \nonumber\\
 \quad &  u\in U \ominus KE  \nonumber\\
 \quad &  \tilde{X}\in X_{set} \ominus E  \nonumber\\
\end{align}
\end{comment}

\section{Stepping Realization on 3D Underactuated Bipedal Robot Cassie}
\label{control}
\begin{figure}[t]
      \centering
      \includegraphics[width = 0.7\columnwidth]{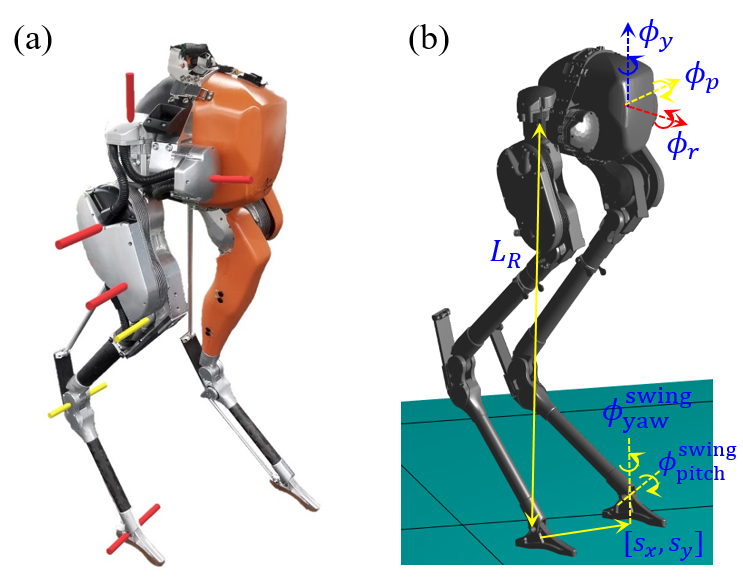}%  inAir.jpg}
      \caption{(a) The physical hardware for the Cassie bipedal robot. (b) An illustration of the output definitions, with the torso orientations, leg length, and swing leg outputs depicted.}
      \label{cassie}
\end{figure}

%%%%%%%%%%%%%%%%%%%%%%%%%%%%%%%%%%%%%%%%%%%%%%%%%%%%%%%%%%%%%%%%%%%%%%%%%%%%%%%%%%%%%%%%%%%%%%%%%%%%%%%%%%
Here we apply the presented H-LIP based stepping on the 3D underactuated bipedal robot Cassie (see Fig. \ref{cassie}). In particular, we apply the planned steps sizes together with the construction of outputs that allow for their realization on the full-order dynamics. 

 The output construction includes three components: the periodic leg length trajectory, the swing foot trajectory, and the orientations of the pelvis and swing foot. The feedback control can be realized via optimization-based controllers, either control Lyapunov function based Quadratic programs (CLF-QPs) \cite{ames2014rapidly, XiongSLIP} or the task space controllers \cite{escande2014hierarchical, wensing2013generation}.

Additionally, Cassie has narrow feet (Fig. \ref{torsoHLIP} (c)), which introduces the underactuated roll motion at the foot contact (underactuated in the lateral plane). We also remove the foot actuation on the stance foot so that the pitch about the ankle is also underactuated (underactuated in the sagittal plane). This fully resembles the point foot nature of the H-LIP.

\subsection{Walking Specification and Construction:}
Walking behaviors are constructed by designing the desired output trajectories of the actuated degrees of freedom. Similar to the approach for generating periodic walking for Cassie \cite{xiong2018coupling, xiong20213d}, the outputs include the leg length, swing foot locations, orientations of the pelvis, and the swing foot.

\textbf{Leg Length:} A periodic time-based leg length trajectory ensures the existence of periodic touch-down and lift-off behavior of the walking. This ensures that the system periodically strikes the ground, and thus the existence of the S2S dynamics of Eq. \eqref{eq:robotS2S}. The vertical height of the COM and the duration of each domain are also approximately constant, matching the assumption on the H-LIP. The desired periodic trajectories of $L^{\text{des}}_{\text{L}}, L^{\text{des}}_{\text{R}}$ can be optimized via an optimization for a stepping-in-place walking motion on the full robot model or on a simple approximated model. Here we apply the approach in \cite{xiong2018coupling} of using an actuated spring-loaded inverted pendulum model to generate the leg length trajectories.

\textbf{Swing Foot Position:} In the SSP, the swing foot position is controlled towards the desired step location to achieve the desired step sizes. The desired step sizes ${u^{x,y}_k}^{\text{des}}$ of the robot are calculated based on the H-LIP stepping in each plane in Eq. \eqref{eq:Hlip-stepping}, which provides
\begin{align}
\label{eq:robotStepsizeX}
{u^x_k}^{\text{des}} &= {u^x_k}^\text{H-LIP} + K_{\text{LQR}}( \mathbf{x}^h_k - \mathbf{x}^\text{H-LIP}_k) \\
\label{eq:robotStepsizeY}
{u^y_k}^{\text{des}} &= {u^y_k}^\text{H-LIP} + K_{\text{LQR}}( \mathbf{y}^h_k - \mathbf{y}^\text{H-LIP}_k) 
\end{align}
where the H-LIP states and step sizes are from the MPC controller in Eq. \eqref{eq:MPC-LIP}. We use the gains from the Linear Quadratic Regulator (LQR) controllers \cite{boyd1991linear} for the linear system to make $(A+BK_{\text{LQR}})$ stable ($A, B$ are from Eq. \eqref{eq:HLIP_S2S}), since tuning the LQR cost can avoid excessive step sizes to be executed on the robot.
%%%% explain the error dynamics here. %%%%%%
%where $K$ is the gain to make $A- BK$ to be a Schur matrix, and $X_k$ is the current state at SSP of the aSLIP. Adding the term $K(X_k - \tilde{X_k})$ not only allows a continuous construction of the step length, but provides a state-dependent feedback term based on the model error. This additional term has existed in \cite{robust MPC literature}.

Unlike the H-LIP, the pre-impact state of the robot cannot be determined before the impact happens. Thus we use the current horizontal state of the robot in SSP to continuously calculate the desired step sizes in Eq. \eqref{eq:robotStepsizeX} \eqref{eq:robotStepsizeY}. The desired swing foot trajectory of the robot will be constructed toward the desired step sizes, which yields horizontal state-dependent trajectories for the swing foot.
Thus the desired swing foot position $\{x,y\}^{\text{des}}_\text{sw}$ in the horizontal $x-y$ plane is:
\begin{align}
\label{eq:stepLengthConstructionX} 
x^{\text{des}}_\text{sw}(t) &=  x_\text{sw}^{+} + c(t)  {u^{x}_k}^{\text{des}}(t), \\
\label{eq:stepLengthConstructionY} 
y^{\text{des}}_\text{sw}(t) & = y_\text{sw}^{+} + c(t) {u^{y}_k}^{\text{des}}(t),
\end{align}
where $c(t)$ is a smooth scalar function which increases from 0 to 1 before $t$ reaches to $T_\text{SSP}$, and $\{x, y\}_\text{sw}^{+}$ is the initial swing foot location in the $x-y$ plane during the SSP.

\textbf{Pelvis and Swing Foot Orientation:} Assuming the robot always turns towards the direction of walking, the desired yaw angle of the pelvis $\phi_{\text{y}}$ is the direction of the path $r^d_\theta (t)$. The desired pitch and roll angles of the pelvis $\phi_{\text{r,p}}$ are set to 0. Thus $\phi^{\text{des}}_{\text{rpy}} = [0,0, r_\theta^d]$. The desired pitch angle of the swing foot is zero. The desired yaw angle of the swing foot $\phi^{\text{pitch}}_{\text{sw}}$ is constructed smoothly from the yaw angle in the beginning of the SSP ${\phi_{\text{sw}}^{\text{yaw}}}^{+}$ to the direction of the path:
 \begin{align}
\label{swingYaw}
{\phi_\text{sw}^\text{yaw}}^{\text{des}}(t)= (1 - c(t)){\phi_{\text{sw}}^{\text{yaw}}}^{+} + c(t) r_\theta^d(t).
\end{align}
Compactly, we can define the outputs in each domain as:
\begin{align*}
& \mathcal{Y}_{\text{DSP}}(q,t) = \begin{bmatrix} L_{\text{L}} , L_{\text{R}}  , \phi_{\text{rpy}}  \end{bmatrix}^T \nonumber - \begin{bmatrix}  L^{\text{des}}_{\text{L}}(t) , L^{\text{des}}_{\text{R}}(t)
, \phi^{\text{des}}_{\text{rpy}}(t)\end{bmatrix}^T \nonumber \\
 &\mathcal{Y}_{\text{SSP}}(q,t) =\begin{bmatrix}    L_{\text{L}} , L_{\text{R}} ,\phi_{\text{rpy}}, x_\text{sw} , y_\text{sw} , \phi^{\text{pitch}}_{\text{sw}}   , \phi^{\text{yaw}}_{\text{sw}}   \end{bmatrix}^T  \nonumber \\
&-
\begin{bmatrix}  L^{\text{des}}_{\text{L}}(t), L^{\text{des}}_{\text{R}}(t) , \phi^{\text{des}}_{\text{rpy}}(t), x^{\text{des}}_\text{sw}(t)  , y^{\text{des}}_\text{sw}(t)
  ,0 , {\phi_\text{sw}^{\text{yaw}}}^{\text{des}}(t) \end{bmatrix}^T. \nonumber
\end{align*}
\begin{comment}
\begin{align}
 \mathbf{y}_{\text{DSP}}(q,t) &= \begin{bmatrix} L_{\text{L}}(q) \\ L_{\text{R}}(q)  \\  \phi_{\text{roll}}(q)   \\ \phi_{\text{pitch}}(q)   \\ \phi_{\text{yaw}}(q)  \\  \end{bmatrix} -
\begin{bmatrix}  L^{\text{des}}_{\text{L}}(t) \\ L^{\text{des}}_{\text{R}}(t)
\\ 0 \\ 0  \\  \theta^d_r(t) \end{bmatrix},
\\
 \mathbf{y}_{\text{SSP}}(q,t) &= \begin{bmatrix} L_{\text{L}}(q) \\ L_{\text{R}}(q)  \\ s_L(q)  \\  s_W(q)   \\ \phi_{\text{roll}}(q)   \\ \phi_{\text{pitch}}(q)   \\ \phi_{\text{yaw}}(q)  \\ \phi_{\text{pitch}}^{\text{swing}}(q)   \\ \phi_{\text{yaw}}^{\text{swing}}(q)   \end{bmatrix} -
\begin{bmatrix}  L^{\text{des}}_{\text{L}}(t) \\ L^{\text{des}}_{\text{R}}(t)  \\   s^{\text{des}}_L(t)  \\  s^{\text{des}}_W(t)
\\ 0  \\ 0  \\ \theta(t)  \\ 0 \\ {\phi_{\text{yaw}}^{\text{swing}}}^{\text{des}}(t) \end{bmatrix}.
\end{align}
\end{comment}
\subsection{Low-level Feedback Control}
The objective of the low-level controller is to stabilize the outputs $\mathcal{Y}_\text{SSP/DSP}$ to zero. Differentiating the outputs twice yield the input torque $\tau$ in output dynamics. A control Lyapunov function (CLF) $V$ can be constructed on the feedback linearized output dynamics \cite{ames2014rapidly}, which provides an inequality constraint on time derivative of $V$, i.e., $\dot{V}$. Thus the exponential stabilization can be enforced by $\dot{V} \leq -\gamma V$ with $\gamma >0$, which is an affine inequality condition on the motor torques $\tau$. A CLF based quadratic program (CLF-QP) can thus be formulated to minimize the input torque subject to the inequality on the CLF, torque limits and ground reaction force constraints. More details can be found in \cite{XiongSLIP, ames2013towards}. Due to the space limit, here we simply denote the CLF-QP controller as:
\begin{equation}
\label{QP}
\tau = \mathcal{U}_{\text{CLF}}(q,\dot{q}, t).
\end{equation}

\begin{algorithm}\caption{H-LIP Stepping for Global Position Control}
 \begin{algorithmic}[1]
 \renewcommand{\algorithmicrequire}{\textbf{Input:}}
 \renewcommand{\algorithmicensure}{\textbf{Output:}}
 \REQUIRE Given desired trajectory $r_x^d(t), r_y^d(t), r_\theta^d(t)$
 %\ENSURE
% \\ \textit{Initialisation} :
\STATE $L^{\text{des}}_{\text{L}}, L^{\text{des}}_{\text{R}}, T_{\text{SSP}}, T_{\text{DSP}} \leftarrow$ Stepping-in-place optimization
%\STATE Initialization of parameters.
\STATE $K_{\text{LQR}}\leftarrow$ Eq. \eqref{eq:HLIP_S2S}
\WHILE {Simulation/Control loop}
\IF {SSP}
   \STATE $x^{\text{des}}_\text{sw}, y^{\text{des}}_\text{sw}$ from Eq. \eqref{eq:stepLengthConstructionX}, \eqref{eq:stepLengthConstructionY}.
       \IF {$\text{SSP}^{-}$}
        \STATE $ {u^x_k}^\text{H-LIP},  {u^y_k}^\text{H-LIP}, \mathbf{x}^\text{H-LIP}_k,\mathbf{y}^\text{H-LIP}_k\leftarrow$ Eq. \eqref{eq:MPC-LIP}
		\ENDIF
%\STATE $\alpha = 0$
\ENDIF
\STATE Update Outputs $\mathcal{Y}_{\text{SSP/DSP}}(q, t)$
\STATE $ \tau\leftarrow$ Eq. \eqref{QP}
\ENDWHILE
 \end{algorithmic}
 \end{algorithm}

\begin{figure}[t]
      \centering
      \includegraphics[width = 1.0\columnwidth]{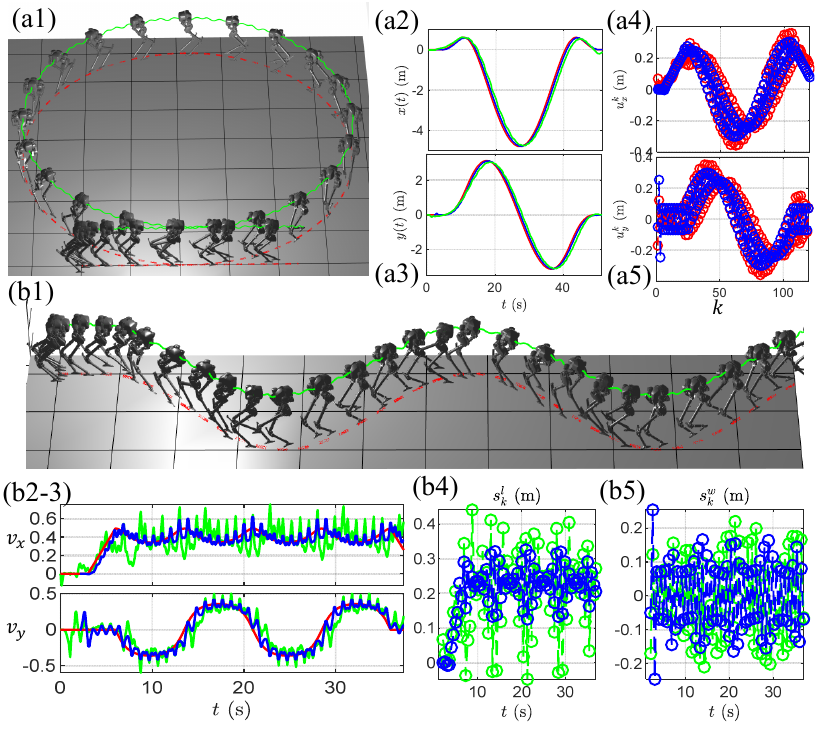}%  inAir.jpg}
      \caption{Simulation results (Red lines are the desired, blue lines are the planned trajectories of the H-LIP, and green lines are the actual trajectories of Cassie): Tracking a circle path (a) and a sinusoidal path (b) in terms of the global position/velocity trajectories (2,3) and the step sizes in the different planes (4, 5).}
      \label{results}
\end{figure}
%
\begin{comment}
\begin{figure*}[t]
      \centering
      \includegraphics[width = 6.3in]{Figures/wideSimV3.PNG}%  inAir.jpg}
      \caption{Simulation results (Red lines are the desired, blue lines are the planned trajectories of the H-LIP, and green lines are the actual trajectories of Cassie): Tracking a circle path (a) and a sinusoidal path (b) in terms of the global position/velocity trajectories (2,3) and the step sizes in the different planes (4, 5).}
      \label{results}
\end{figure*}
\end{comment}

\section{Results and Discussion}
In this section, we present the results of implementing the proposed approach on the robot Cassie to realize several tasks under the global position control. A video of the results can be seen in \cite{Supplementary}.

\subsection{Setup} \label{sec:simulation_setup}
The proposed method of step planning for global position control is summarized in Algorithm 1. In particular, we evaluate the method on Cassie in simulation. The dynamics is integrated using Matlab \texttt{ODE 45} function with event-based triggering for contact and domain switching. The MPC formulated QP is solved at the impact event, and CLF-QP is solved at 1kHz using qpOASES \cite{Ferreau2014}. The disturbance invariant set is calculated using MPT3 \cite{MPT3}.
%For the experimental results, we consider a single desired path consisting of a square.  This will be further discussed after describing the simulation results.

We designed various shapes of paths, including a circle, a cardioid, a square, and a sinusoid, for the robot to track. %A walking speed profile is customized for tracking the path. 
Trapezoidal speed profiles are designed for tracking the cardioid, the circle, and the sinusoid path. A triangle speed profile is used on the square path. For the experimental results, we consider the square path. This will be further discussed after describing the simulation results. 
\subsection{Path Tracking}
Using the method in this paper, the robot can track all paths well in terms of position and velocity profiles (Fig. \ref{results}). The circle and cardioid can be tracked easily even at relatively large speeds. The sinusoid and square paths are designed as challenging examples. On the sinusoid, the robot walks and turns significantly. While on the square, the path is not smooth, thus the robot has to come to a stop and turn. More dynamical maneuvers, such as sharp turning with high speeds, are challenging to realize it on the robot considering the existence of the kinematic limits and torque bounds. 

Undoubtedly, the tracking can fail when the required walking and turning speeds are too large, e.g., over 1m/s and 45deg/s, at the same time. The failure often happens when the step sizes are too large for the robot to track due to limited motor torques. This can be easily avoided by setting limits on the allowed forward and turning velocities. 

\begin{figure}[t]
      \centering
      \includegraphics[width = 3.4in]{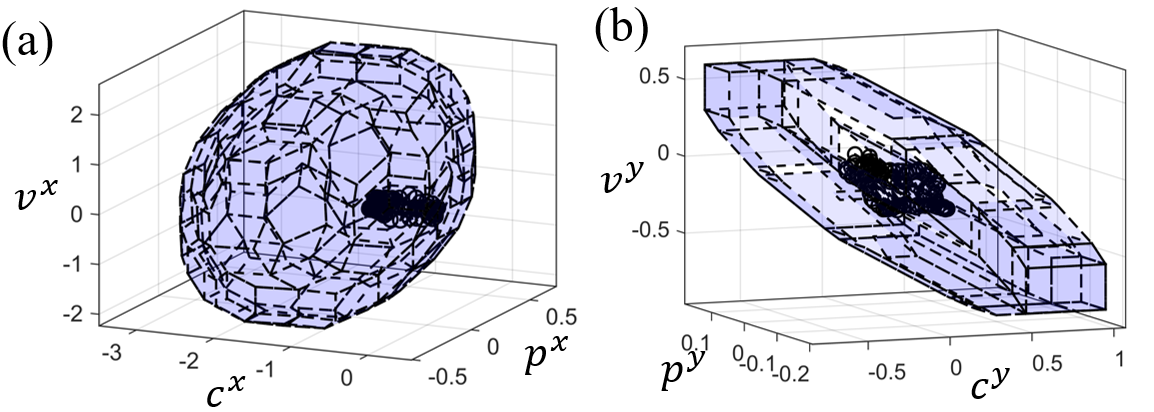}
      \caption{The tracking error $\mathbf{e}$ (black circles) between the robot states and the H-LIP plotted with the disturbance invariant set $E$ (transparent polytopes) for the case of tracking the cardioid path in the sagittal (a) and lateral (b) plane. }
      \label{set}
\end{figure}

The tracking error is analyzed by the disturbance invariant set $E$. The model difference $w$ between the S2S dynamics of the robot and that of the H-LIP is numerically calculated in simulation. By bounding the error by a polytopic set $W$, we approximate $E$ by a polytope. All the tracking errors are indeed inside the approximation of the set $E$. This is shown in Fig. \ref{set}, which verifies our approach. 

\subsection{Additional Verification}

The path tracking is considered as the basic verification of the proposed approach. In the following, we present more challenging control problems that are solved using the stepping method. 

\begin{figure}[t]
      \centering
      \includegraphics[width = 1.0\columnwidth]{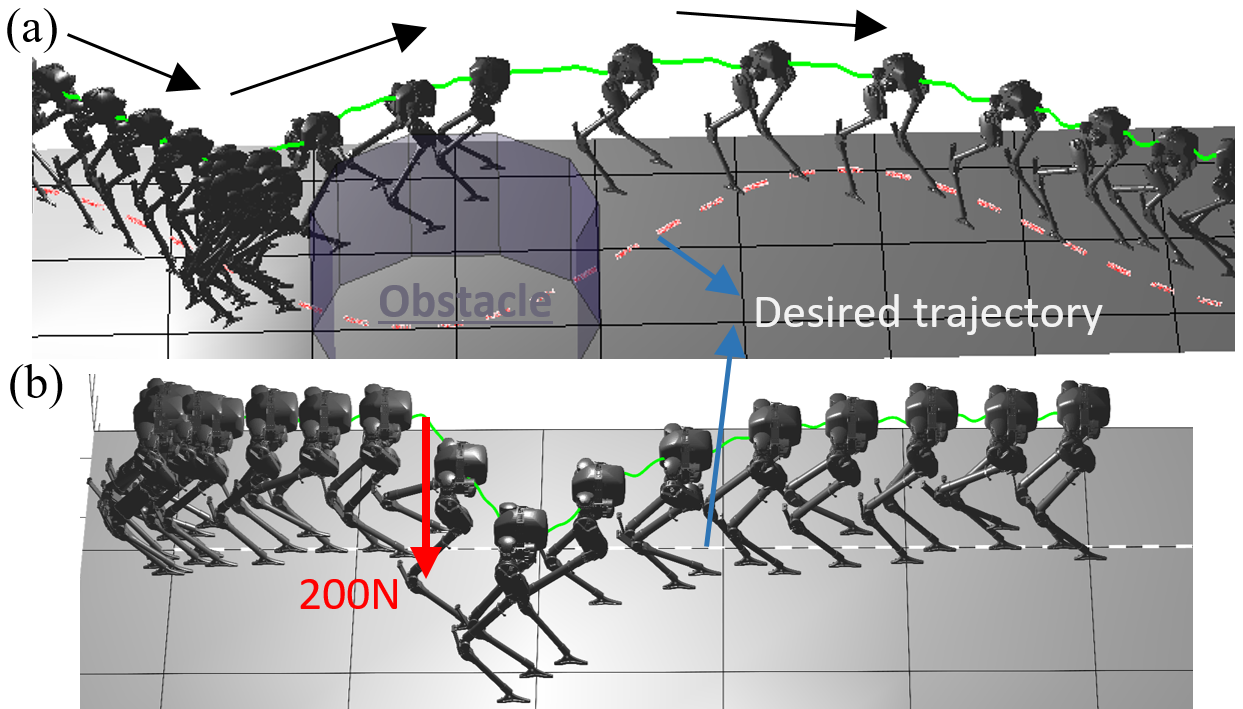}%  inAir.jpg}
      \caption{The simulated walking with (a) avoiding an obstacle on the path and (b) disturbance rejection.}
      \label{fig:obstacle}
\end{figure}

\noindent{\textbf{Obstacle Avoidance:}} The given path should be obstacle-free. In the case of existing obstacles on the path, we can avoid obstacles by adding constraints in the MPC formulation, e.g., in the form of 
\begin{equation}
|\mathbf{c}(t) - \mathbf{p}_{\text{obstacle}} | > d,
\end{equation}
where $\mathbf{p}_{\text{obstacle}}$ is the position of the obstacle and $d$ is the distance to keep away from the it. This changes the MPC from a QP into a quadratic constrained quadratic program (QCQP), which is also fast to solve on the low dimensional system. Fig. \ref{fig:obstacle} (a) shows an example of avoiding a circular obstacle while tracking a sinusoidal path. After avoiding the obstacle, the robot walks back on the path. 

\noindent{\textbf{Disturbance Rejection:}} The step planning approach can also handle unknown external disturbances on the robot during walking. Fig. \ref{fig:obstacle} (b) shows an example where an external lateral force of $200$N is added on the pelvis at $t = 10$s for $0.1$s duration. The robot was pushed away from the path but then walked back to its original path.  

\noindent{\textbf{Experiment:}} The tracking for a square path was implemented on Cassie as a preliminary experiment to demonstrate that the generated motion from simulation using this approach are dynamically achievable on hardware. The experiment used the simulated walking velocities and turning rates of the robot as targets, which were tracked on the robot with an existing controller described in \cite{reher2020inverse}. 
 \begin{figure}[t]
      \centering
      \includegraphics[width = 1.0\columnwidth]{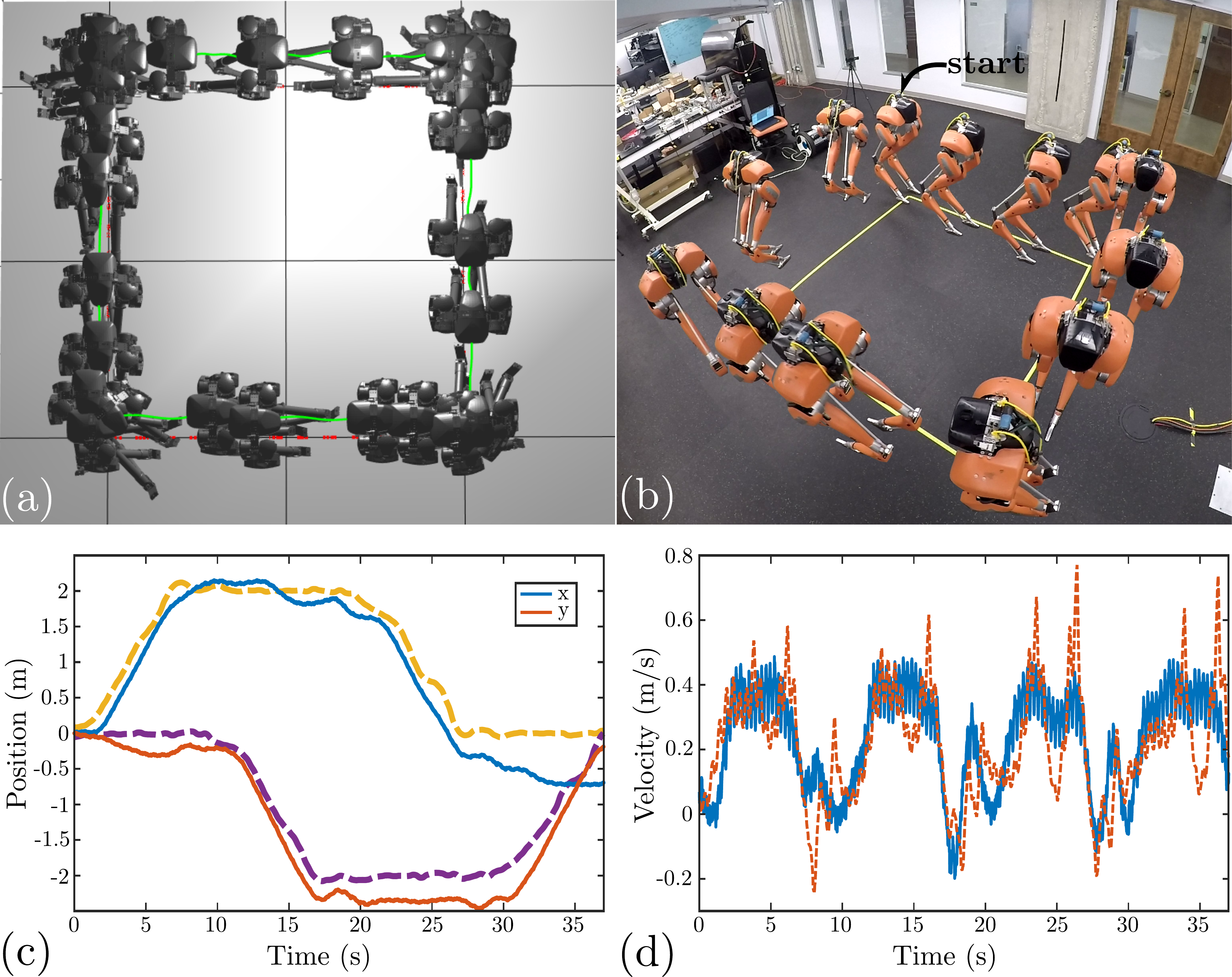}%  inAir.jpg}
      \caption{Cassie follows the square path in simulation (a) and in experiment (b). The pelvis global positions (c) and heading velocities (d) from simulation (dashed) and the experiment (solid).}
      \label{fig:experiment}
\end{figure}

The robot ultimately followed the desired walking profiles on hardware, with a comparison of the positions and velocities of the experiment and simulation shown in Fig. \ref{fig:experiment}. As the robot is tracking the commanded velocities, there is a slight drift in the position and heading as it executes the later segments of the shape. The methods proposed in this work will ultimately be realized on hardware in future work to provide a holistic experimental control approach.

%Because path tracking requires high fidelity global estimation of the robotic pose, it is not directly 
%We are in the progress of realizing the proposed approach on the hardware. The path tracking requires the state estimation including the global position of the robot, for which we are trying using a camera. It is still in progress in the lab and requires significant amount of work. To demonstrate the approach in this paper at this moment, we realized the simulated walking and turning velocities on the robot without the global position feedback loop. The robot does not track the path ideally, but this shows promise to realize the approach when the global position feedback is available. 

%\textbf{Discussion:} The proposed approach only requires to solve linear MPC (QP) at discrete step-level for planning and QPs for local joint control. Thus it requires little computation for realizing the path tracking. However, at this moment, kinematic collision avoidance is still an unsolved problem, especially for the robot Cassie which has complicated leg design. Then the robot is being pushed, The   Additionally, the speeds of the    Kinematic collision avoidance. 
% speed of tracking
%%%%%%%%%%%%%%%%%%%%%%%%%%%%%%%%%%%%%%%%%%%%%%%%%%%%%%%%%%%%%%%%%%%%%%%%%%%%%%%%%%%%%%%%%%%%%%%%%%%%%%%
\section{Conclusion and Future Work}
Global position control on 3D underactuated walking robots is enabled through the footstep planning on the Hybrid Linear Inverted Pendulum (H-LIP) model. Simulation of the controller on the robot Cassie demonstrates the proposed method. %The main idea is to use the step-to-step dynamics of the H-LIP as the approximation of the translational dynamics of the 3D robot during walking. \
The procedures of realizing this controller can be viewed at three levels: the first is the MPC on the H-LIP; the second is the feedback adjustment based on the H-LIP stepping; the last is the local output stabilization via the CLF-QPs.

Future work will be on the improvement of the approximation via numerical methods for reducing $w$ in Eq. \eqref{eq:w}. Systematic construction of the output of the walking on the robot can also be improved to make the robot closer to the walking on the H-LIP. Robust optimal feedback controllers can also be developed to minimize the disturbance invariant set $E$, which can reduce tracking error between the approximation and the full robot model. More importantly, we would like to work towards a formal realization of the full approach on the hardware, including the global position feedback. The hope is that this enables dynamic walking on underactuated robots autonomously in dynamic and challenging environments. %Analysis of the model difference (disturbance to the approximation) can help find optimal feedback gain $K$ to reduce the size of the invariant set of the closed-loop system in Eq. \ref{closedLoopSystem} to reduce tracking errors.

%%%% WHEN THERE IS FOOT ACTUATION. best and rigorously apply the available foot actuation.
% \section*{Acknowledgments}

%% Use plainnat to work nicely with natbib. 
%\addtolength{\textheight}{-6.3cm}
\addtolength{\textheight}{0cm}  
\bibliographystyle{IEEEtran}
\bibliography{references}

\begin{thebibliography}{10}
\providecommand{\url}[1]{#1}
\csname url@samestyle\endcsname
\providecommand{\newblock}{\relax}
\providecommand{\bibinfo}[2]{#2}
\providecommand{\BIBentrySTDinterwordspacing}{\spaceskip=0pt\relax}
\providecommand{\BIBentryALTinterwordstretchfactor}{4}
\providecommand{\BIBentryALTinterwordspacing}{\spaceskip=\fontdimen2\font plus
\BIBentryALTinterwordstretchfactor\fontdimen3\font minus
  \fontdimen4\font\relax}
\providecommand{\BIBforeignlanguage}[2]{{%
\expandafter\ifx\csname l@#1\endcsname\relax
\typeout{** WARNING: IEEEtran.bst: No hyphenation pattern has been}%
\typeout{** loaded for the language `#1'. Using the pattern for}%
\typeout{** the default language instead.}%
\else
\language=\csname l@#1\endcsname
\fi
#2}}
\providecommand{\BIBdecl}{\relax}
\BIBdecl

\bibitem{clary2018monte}
P.~Clary, P.~Morais, A.~Fern, and J.~Hurst,
  ``\href{https://www.aaai.org/ocs/index.php/ICAPS/ICAPS18/paper/download/17789/16936
  }{Monte-Carlo Planning for Agile Legged Locomotion},'' in \emph{Twenty-Eighth
  International Conference on Automated Planning and Scheduling}, 2018.

\bibitem{agrawal2017discrete}
A.~Agrawal and K.~Sreenath,
  ``\href{http://www.roboticsproceedings.org/rss13/p73.pdf }{Discrete Control
  Barrier Functions for Safety-Critical Control of Discrete Systems with
  Application to Bipedal Robot Navigation.}'' in \emph{Robotics: Science and
  Systems}, 2017.

\bibitem{scianca2019mpc}
N.~Scianca, D.~De~Simone, L.~Lanari, and G.~Oriolo,
  ``\href{https://arxiv.org/abs/1901.08505 }{MPC for Humanoid Gait Generation:
  Stability and Feasibility},'' \emph{arXiv preprint arXiv:1901.08505}, 2019.

\bibitem{apgar2018fast}
T.~Apgar, P.~Clary, K.~Green, A.~Fern, and J.~W. Hurst, ``Fast online
  trajectory optimization for the bipedal robot cassie.'' in \emph{Robotics:
  Science and Systems}, vol. 101, 2018, p.~14.

\bibitem{4115592}
P.~{Wieber}, ``Trajectory free linear model predictive control for stable
  walking in the presence of strong perturbations,'' in \emph{2006 6th IEEE-RAS
  International Conference on Humanoid Robots}, pp. 137--142.

\bibitem{8794117}
A.~{Pajon} and P.~{Wieber}, ``Safe 3d bipedal walking through linear mpc with
  3d capturability,'' in \emph{2019 International Conference on Robotics and
  Automation (ICRA)}, pp. 1404--1409.

\bibitem{naveau2016reactive}
M.~Naveau, M.~Kudruss, O.~Stasse, C.~Kirches, K.~Mombaur, and P.~Sou{\`e}res,
  ``\href{https://ieeexplore.ieee.org/document/7384453 }{A reactive walking
  pattern generator based on nonlinear model predictive control},'' \emph{IEEE
  Robotics and Automation Letters}, vol.~2, no.~1, pp. 10--17, 2016.

\bibitem{fallon2015architecture}
M.~Fallon, S.~Kuindersma, S.~Karumanchi, M.~Antone, T.~Schneider, H.~Dai, C.~P.
  D'Arpino, R.~Deits, M.~DiCicco, D.~Fourie \emph{et~al.}, ``An architecture
  for online affordance-based perception and whole-body planning,''
  \emph{Journal of Field Robotics}, vol.~32, no.~2, pp. 229--254, 2015.

\bibitem{tanguy2019closed}
A.~Tanguy, D.~De~Simone, A.~I. Comport, G.~Oriolo, and A.~Kheddar,
  ``\href{https://ieeexplore.ieee.org/document/8794006 }{Closed-loop MPC with
  Dense Visual SLAM-Stability through Reactive Stepping},'' in \emph{2019
  International Conference on Robotics and Automation (ICRA)}.\hskip 1em plus
  0.5em minus 0.4em\relax IEEE, 2019, pp. 1397--1403.

\bibitem{vukobratovic2004zero}
M.~Vukobratovi{\'c} and B.~Borovac,
  ``\href{https://www.worldscientific.com/doi/10.1142/S0219843604000083
  }{Zero-moment point—thirty five years of its life},'' \emph{International
  journal of humanoid robotics}, vol.~1, no.~01, pp. 157--173, 2004.

\bibitem{kajita2003biped}
S.~Kajita, F.~Kanehiro, K.~Kaneko, K.~Fujiwara, K.~Harada, K.~Yokoi, and
  H.~Hirukawa, ``\href{https://ieeexplore.ieee.org/document/1241826 }{Biped
  walking pattern generation by using preview control of zero-moment point},''
  in \emph{Robotics and Automation, IEEE International Conference on}, vol.~2,
  2003, pp. 1620--1626.

\bibitem{8815144}
Y.~{Gao} and Y.~{Gu}, ``Global-position tracking control of a fully actuated
  nao bipedal walking robot,'' in \emph{2019 American Control Conference
  (ACC)}, 2019, pp. 4596--4601.

\bibitem{8461140}
T.~{Seyde}, A.~{Shrivastava}, J.~{Englsberger}, S.~{Bertrand}, J.~{Pratt}, and
  R.~J. {Griffin}, ``Inclusion of angular momentum during planning for capture
  point based walking,'' in \emph{2018 IEEE International Conference on
  Robotics and Automation (ICRA)}, 2018, pp. 1791--1798.

\bibitem{rezazadeh2015spring}
S.~Rezazadeh and et~al.,
  ``\href{https://proceedings.asmedigitalcollection.asme.org/DSCC/proceedings-abstract/DSCC2015/57243/V001T04A003/228027
  }{Spring-mass walking with atrias in 3d: Robust gait control spanning zero to
  4.3 kph on a heavily underactuated bipedal robot},'' in \emph{ASME 2015
  dynamic systems and control conference}.\hskip 1em plus 0.5em minus
  0.4em\relax American Society of Mechanical Engineers, 2015.

\bibitem{raibert1986legged}
M.~H. Raibert, \emph{\href{https://mitpress.mit.edu/books/legged-robots-balance
  }{Legged robots that balance}}.\hskip 1em plus 0.5em minus 0.4em\relax MIT
  press, 1986.

\bibitem{SreenathPPG11}
K.~Sreenath, H.~Park, I.~Poulakakis, and J.~W. Grizzle,
  ``\href{https://hybrid-robotics.berkeley.edu/publications/IJRR2011.pdf }{A
  Compliant Hybrid Zero Dynamics Controller for Stable, Efficient and Fast
  Bipedal Walking on {MABEL}},'' \emph{I. J. Robotics Res.}, vol.~30, no.~9,
  pp. 1170--1193, 2011.

\bibitem{da20162d}
X.~Da, O.~Harib, R.~Hartley, B.~Griffin, and J.~Grizzle,
  ``\href{https://ieeexplore.ieee.org/document/7501826 }{From 2D design of
  underactuated bipedal gaits to 3D implementation: Walking with speed
  tracking},'' \emph{IEEE Access}, vol.~4, pp. 3469--3478, 2016.

\bibitem{xin2019online}
S.~Xin, R.~Orsolino, and N.~G. Tsagarakis, ``Online relative footstep
  optimization for legged robots dynamic walking using discrete-time model
  predictive control.'' in \emph{IROS}, 2019, pp. 513--520.

\bibitem{hodgins1991adjusting}
J.~K. Hodgins and M.~Raibert,
  ``\href{https://ieeexplore.ieee.org/stamp/stamp.jsp?arnumber=88138
  }{Adjusting step length for rough terrain locomotion},'' \emph{IEEE
  Transactions on Robotics and Automation}, vol.~7, no.~3, pp. 289--298, 1991.

\bibitem{kim2020dynamic}
D.~Kim, S.~J. Jorgensen, J.~Lee, J.~Ahn, J.~Luo, and L.~Sentis, ``Dynamic
  locomotion for passive-ankle biped robots and humanoids using whole-body
  locomotion control,'' \emph{The International Journal of Robotics Research},
  vol.~39, no.~8, pp. 936--956, 2020.

\bibitem{westervelt2003hybrid}
E.~R. Westervelt, J.~W. Grizzle, and D.~E. Koditschek,
  ``\href{https://ieeexplore.ieee.org/stamp/stamp.jsp?arnumber=1166523 }{Hybrid
  zero dynamics of planar biped walkers},'' \emph{IEEE transactions on
  automatic control}, vol.~48, no.~1, pp. 42--56, 2003.

\bibitem{garcia1998simplest}
M.~Garcia, A.~Chatterjee, A.~Ruina, and M.~Coleman,
  ``\href{https://www.ncbi.nlm.nih.gov/pubmed/10412391 }{The simplest walking
  model: stability, complexity, and scaling},'' \emph{Journal of biomechanical
  engineering}, vol. 120, no.~2, pp. 281--288, 1998.

\bibitem{powell2015model}
M.~J. Powell, E.~A. Cousineau, and A.~D. Ames,
  ``\href{https://ieeexplore.ieee.org/abstract/document/7139912/ }{Model
  predictive control of underactuated bipedal robotic walking},'' in \emph{2015
  IEEE International Conference on Robotics and Automation (ICRA)}.\hskip 1em
  plus 0.5em minus 0.4em\relax IEEE, 2015, pp. 5121--5126.

\bibitem{chen2020optimal}
Y.-M. Chen and M.~Posa, ``Optimal reduced-order modeling of bipedal
  locomotion,'' in \emph{2020 IEEE International Conference on Robotics and
  Automation (ICRA)}, pp. 8753--8760.

\bibitem{nguyen2017dynamic}
Q.~Nguyen, A.~Agrawal, X.~Da, W.~C. Martin, H.~Geyer, J.~W. Grizzle, and
  K.~Sreenath, ``\href{http://www.roboticsproceedings.org/rss13/p72.pdf
  }{Dynamic Walking on Randomly-Varying Discrete Terrain with One-step
  Preview.}'' in \emph{Robotics: Science and Systems}, 2017.

\bibitem{xiong20213d}
X.~Xiong and A.~Ames, ``3d underactuated bipedal walking via h-lip based gait
  synthesis and stepping stabilization,'' \emph{arXiv preprint
  arXiv:2101.09588}, 2021.

\bibitem{xiong2020ral}
X.~{Xiong} and A.~D. {Ames}, ``Dynamic and versatile humanoid walking via
  embedding 3d actuated slip model with hybrid lip based stepping,'' \emph{IEEE
  Robotics and Automation Letters}, vol.~5, no.~4, pp. 6286--6293, 2020.

\bibitem{krause2012stabilization}
M.~Krause, J.~Englsberger, P.-B. Wieber, and C.~Ott,
  ``\href{https://www.sciencedirect.com/science/article/pii/S1474667016336059
  }{Stabilization of the capture point dynamics for bipedal walking based on
  model predictive control},'' \emph{IFAC Proceedings Volumes}, vol.~45,
  no.~22, pp. 165--171, 2012.

\bibitem{xiong2021slip}
X.~Xiong and A.~Ames, ``Slip walking over rough terrain via h-lip stepping and
  backstepping-barrier function inspired quadratic program,'' \emph{IEEE
  Robotics and Automation Letters}, 2021.

\bibitem{faraji2014robust}
S.~Faraji, S.~Pouya, and A.~Ijspeert,
  ``\href{https://infoscience.epfl.ch/record/198512?ln=en }{Robust and agile 3d
  biped walking with steering capability using a footstep predictive
  approach},'' in \emph{Robotics science and systems (RSS)}, no. CONF, 2014.

\bibitem{villa2017model}
N.~A. Villa and P.-B. Wieber, ``Model predictive control of biped walking with
  bounded uncertainties,'' in \emph{2017 IEEE-RAS 17th International Conference
  on Humanoid Robotics (Humanoids)}.\hskip 1em plus 0.5em minus 0.4em\relax
  IEEE, 2017, pp. 836--841.

\bibitem{griffin2018straight}
R.~J. Griffin, G.~Wiedebach, S.~Bertrand, A.~Leonessa, and J.~Pratt,
  ``Straight-leg walking through underconstrained whole-body control,'' in
  \emph{2018 IEEE International Conference on Robotics and Automation
  (ICRA)}.\hskip 1em plus 0.5em minus 0.4em\relax IEEE, 2018, pp. 1--5.

\bibitem{ames2014rapidly}
A.~D. Ames, K.~Galloway, K.~Sreenath, and J.~Grizzle,
  ``\href{https://ieeexplore.ieee.org/stamp/stamp.jsp?arnumber=6709752
  }{Rapidly exponentially stabilizing control lyapunov functions and hybrid
  zero dynamics},'' \emph{IEEE Transactions on Automatic Control}, vol.~59,
  no.~4, pp. 876--891, 2014.

\bibitem{XiongSLIP}
X.~Xiong and A.~D. Ames, ``Bipedal hopping: Reduced-order model embedding via
  optimization-based control,'' in \emph{2018 IEEE/RSJ International Conference
  on Intelligent Robots and Systems (IROS)}, pp. 3821--3828.

\bibitem{escande2014hierarchical}
A.~Escande, N.~Mansard, and P.-B. Wieber,
  ``\href{https://hal.archives-ouvertes.fr/hal-00751924/document }{Hierarchical
  quadratic programming: Fast online humanoid-robot motion generation},''
  \emph{The International Journal of Robotics Research}, vol.~33, no.~7, pp.
  1006--1028, 2014.

\bibitem{wensing2013generation}
P.~M. Wensing and D.~E. Orin,
  ``\href{https://ieeexplore.ieee.org/document/6631008 }{Generation of dynamic
  humanoid behaviors through task-space control with conic optimization},'' in
  \emph{2013 IEEE International Conference on Robotics and Automation}.\hskip
  1em plus 0.5em minus 0.4em\relax IEEE, 2013, pp. 3103--3109.

\bibitem{xiong2018coupling}
X.~Xiong and A.~D. Ames,
  ``\href{https://ieeexplore.ieee.org/stamp/stamp.jsp?arnumber=8625066}{Coupling
  reduced order models via feedback control for 3d underactuated bipedal
  robotic walking},'' in \emph{2018 IEEE-RAS 18th International Conference on
  Humanoid Robots (Humanoids)}, pp. 1--9.

\bibitem{boyd1991linear}
S.~P. Boyd and C.~H. Barratt,
  \emph{\href{https://web.stanford.edu/~boyd/lcdbook/lcdbook.pdf }{Linear
  controller design: limits of performance}}.\hskip 1em plus 0.5em minus
  0.4em\relax Prentice Hall Englewood Cliffs, NJ, 1991.

\bibitem{ames2013towards}
A.~D. Ames and M.~Powell, ``\href{https://arxiv.org/abs/1910.00684 }{Towards
  the unification of locomotion and manipulation through control lyapunov
  functions and quadratic programs},'' in \emph{Control of Cyber-Physical
  Systems}.\hskip 1em plus 0.5em minus 0.4em\relax Springer, 2013, pp.
  219--240.

\bibitem{Supplementary}
Results: \url{https://youtu.be/06efo-U1mrw}.

\bibitem{Ferreau2014}
H.~Ferreau, C.~Kirches, A.~Potschka, H.~G. Bock, and M.~Diehl,
  ``\href{https://link.springer.com/article/10.1007/s12532-014-0071-1
  }{{qpOASES}: A parametric active-set algorithm for quadratic programming},''
  \emph{Mathematical Programming Computation}, vol.~6, no.~4, pp. 327--363,
  2014.

\bibitem{MPT3}
M.~Herceg, M.~Kvasnica, C.~N. Jones, and M.~Morari,
  ``\href{https://ieeexplore.ieee.org/document/6669862 }{{Multi-Parametric
  Toolbox 3.0}},'' in \emph{Proc.~of the European Control Conference},
  Z\"urich, Switzerland, July 17--19 2013, pp. 502--510,
  \url{http://control.ee.ethz.ch/~mpt}.

\bibitem{reher2020inverse}
J.~Reher and A.~D. Ames, ``Inverse dynamics control of compliant hybrid zero
  dynamic walking,'' \emph{arXiv preprint arXiv:2010.09047}, 2020.

\end{thebibliography}

\end{document}